\documentclass[runningheads]{llncs}
\usepackage[T1]{fontenc}
\usepackage{subfiles}

\usepackage{graphicx}

\begin{document}

\title{Assessing Foundation Models' Transferability to Physiological Signals in Precision Medicine}

\titlerunning{Assessing Foundation Models' Transferability for Physiological Signals}

\author{Matthias Christenson\inst{1,6}\orcidID{0000-0003-4858-6541} \and
Cove Geary\inst{1,2}\orcidID{0009-0004-3363-9086} \and
Brian Locke\inst{1,3,4}\orcidID{0000-0002-3588-5238} \and
Pranav Koirala\inst{1,5}\orcidID{0000-0002-5814-9786} \and
Warren Woodrich Pettine\inst{1,6}\orcidID{0000-0003-0063-0902}}
\authorrunning{Assessing Foundation Models' Transferability for Physiological Signals}

\institute{Mountain Biometrics Inc, Salt Lake City UT \and
Department of Psychological and Brain Sciences, Dartmouth College, Hanover NH \and
Intermountain Health, Department of Pulmonary and Critical Care, Murray UT \and
University of Utah, Division of Pulmonary, Critical Care, and Occupational Pulmonary Medicine, Salt Lake City UT
\and
Department of Emergency Medicine, University of Maryland, Baltimore MD \and
Huntsman Mental Health Institute, Department of Psychiatry, University of Utah, Salt Lake City UT\\
Correspondence should be directed to \email{warren@mountainbiometrics.com}}
\maketitle              
\begin{abstract}

The success of precision medicine requires computational models that can effectively process and interpret diverse physiological signals across heterogeneous patient populations. While foundation models have demonstrated remarkable transfer capabilities across various domains, their effectiveness in handling individual-specific physiological signals - crucial for precision medicine - remains largely unexplored. This work introduces a systematic pipeline for rapidly and efficiently evaluating foundation models' transfer capabilities in medical contexts, addressing three key questions: (1) How well do foundation models transfer to precision medicine applications? (2) Can we assess their transfer capacity efficiently? (3) How can we optimize these assessments to guide targeted fine-tuning?

Our pipeline employs a three-stage approach. First, it leverages physiological simulation software to generate diverse, clinically relevant scenarios, particularly focusing on data-scarce medical conditions. This simulation-based approach enables both targeted capability assessment and subsequent model fine-tuning. Second, the pipeline projects these simulated signals through the foundation model to obtain embeddings, which are then evaluated using linear methods. This evaluation quantifies the model's ability to capture three critical aspects: physiological feature independence, temporal dynamics preservation, and medical scenario differentiation. Finally, the pipeline validates these representations through specific downstream medical tasks.

Initial testing of our pipeline on the Moirai time series foundation model revealed significant limitations in physiological signal processing, including feature entanglement, temporal dynamics distortion, and reduced scenario discrimination. These findings suggest that current foundation models may require substantial architectural modifications or targeted fine-tuning before deployment in clinical settings.

Our ongoing work focuses on three primary directions: (1) expanding our simulation framework to encompass a broader range of precision medicine scenarios, particularly those involving patient-specific variations in treatment response; (2) incorporating additional validation tasks that directly assess clinical utility, such as risk stratification and early warning detection; and (3) extending our evaluation to multiple foundation model architectures to identify optimal approaches for medical applications. This comprehensive evaluation framework represents a crucial step toward developing more reliable and clinically relevant foundation models for precision medicine.

\end{abstract}
\section{Introduction}

The success of contemporary artificial intelligence (AI) methods in domains such as language stands in stark contrast to the limited progress in areas of precision medicine involving physiological signals, such as trauma care or remote patient monitoring. This disparity primarily stems from the fundamental challenge of data availability \cite{adadi_survey_2021}. While language models benefit from vast amounts of diverse text data available through books and the internet - forcing them to learn flexible distributions over possible meanings - medical datasets containing physiological signals are comparatively scarce. This scarcity has confined AI models in trauma care to narrow applications, such as hemorrhagic shock \cite{peng_artificial_2023} or sepsis \cite{teng_review_2020}, often providing generic rather than patient-specific responses. Although precision AI in medicine is advancing rapidly, progress remains largely concentrated in data-rich areas, such as electronic medical records and medical imaging \cite{li_neural_2022,barragan-montero_artificial_2021,saab_capabilities_2024}. Without significant innovation in training approaches for data-scarce medical scenarios, the advancement of precision medicine in physiological monitoring will remain limited.

To address data scarcity challenges \cite{adadi_survey_2021}, researchers have developed several approaches, with two being particularly promising for precision medicine applications. The first approach, "data augmentation", artificially expands existing datasets through various transformations - similar to how vision models use techniques like cropping or rotation to create additional training examples. The second approach, "transfer learning", leverages knowledge gained from data-rich domains to improve performance in data-scarce scenarios. For instance, a vision model initially trained on internet images might be fine-tuned for radiological applications. While these techniques have existed for years, recent advances in both areas have created unprecedented opportunities for addressing data-scarce medical problems involving physiological data.

The evolution of transfer learning has been dramatically accelerated by foundation models \cite{bommasani_opportunities_2022}, which have demonstrated unprecedented generalization capabilities across diverse domains - from maze navigation \cite{yu_l3mvn_2023} and genetic analysis \cite{consens_transformers_2023} to time series forecasting \cite{goswami_moment_2024} and beyond \cite{bubeck_sparks_2023}. An increasingly common approach involves: 1) engineering domain-specific input layers to process specialized data types; 2) utilizing existing or minimally-trained foundation models to generate rich embeddings; and 3) training task-specific output layers using these embeddings. This architecture is particularly valuable for precision medicine, as it enables the integration of embeddings from multiple data modalities - such as biophysical signals, genetic information, and clinical history - to deliver truly personalized care. However, understanding the limitations and potential of foundation models requires extensive empirical investigation \cite{goodfellow_deep_2016}, making domain-specific benchmarking crucial for medical applications.

In response to these challenges and opportunities, we present a work-in-progress pipeline designed for rapid, cost-effective, and targeted assessment of foundation models' transfer learning capabilities for physiological signals in precision medicine. Our pipeline uniquely integrates physiological simulations. We demonstrate the pipeline's utility by applying it to the recently released Moirai time series foundation model \cite{woo_unified_2024}, revealing the necessity of extensive fine-tuning for physiological signal processing. Our ongoing work focuses on expanding the range of precision medicine scenarios and validation tasks, while applying the pipeline to evaluate a comprehensive set of time series foundation models \cite{liang_foundation_2024}. 
This work contributes to the ongoing effort to adapt modern AI capabilities for physiological signal processing in precision medicine applications.

\section{Methods}

\subsection{Creation of synthetic physiological signals for precision medicine.}
The BioGears simulation package \cite{baird_biogears_2020} was used to generate synthetic data relevant for precision medicine scenarios in intensive care, trauma and treatment responsiveness. For our initial development of the pipeline, we relied on the default scenarios provided with the package, including acute hemorrhage, sepsis, and multi-organ failure. Each scenario was simulated across multiple virtual patients with varying demographic characteristics to ensure diversity in the generated signals. 
Since each scenario had a different number of physiological signals and different types of signals, we focused our analysis on the seven most common features: arterial pressure, carbon dioxide production rate, central venous pressure, heart rate, oxygen consumption rate, renal blood flow, and respiration rate (Fig. \ref{fig:pipeline_diagram}). Signals were normalized to zero mean and unit variance within each scenario.

\subsection{Projection into the foundation model embeddings.}
The simulated time series data is first reformatted using linear interpolation of values such that it is equal length across scenarios, with each sequence standardized to 1000 timesteps. Each feature for each scenario is independently passed to the transformer module of the foundation model under assessment, maintaining the temporal ordering of the signals. The final embedding layer of the module is used for further analyses. For pipeline development, we used the Moirai time series foundation model \cite{woo_unified_2024}, which was pre-trained on a diverse set of time series data from multiple domains. No fine-tuning was performed prior to our analyses to assess the model's zero-shot transfer capabilities.

\subsection{Assessing embeddings' representation of features, temporal dynamics and scenarios, then validating with a task.} 

\textbf{Feature correlations.}
The pipeline first computes the Pearson correlation coefficient between all pairs of features within each scenario. It then calculates the mean correlation across features and scenarios to quantify the degree of feature entanglement in the embeddings. The ability of the model to reconstruct the raw time series from the embedded time series is assessed using linear regression with 5-fold cross-validation, summarized with the test R$^2$. For any regression or classification task, the dataset is split into train and test sets (proportions of 0.8 and 0.2) with stratification by scenario type. Embeddings are assessed both for reconstructing their own time series as well as the time series of other features to evaluate cross-feature information leakage.

\textbf{Temporal dynamics.}
The pipeline assesses low-dimensional dynamics by performing principal component analysis (PCA) within each scenario across features, both for the raw signals and the embeddings. We retain enough components to explain 90\% of the variance in each case. The pipeline then calculates the average dimensionality of the raw and embedded trajectories, as well as the smoothness of the low-dimensional trajectories. Smoothness of the trajectory is measured by first calculating the mean differences (velocity) across subsequent samples within each scenario normalized by the magnitude of the trajectory. This value is divided by the mean velocity of 1000 randomly permuted trajectories and then subtracted from 1 to obtain a smoothness metric between 0 and 1. A value of 0 corresponds to random trajectories and 1 corresponds to smooth trajectories. This analysis reveals whether the model preserves the temporal structure inherent in physiological signals.

\textbf{Scenario correlations.}
To characterize the model's capacity for distinguishing medical scenarios, the pipeline compares the correlation between scenarios for both the raw signal and embedding using cosine similarity. As a summary metric, it uses the mean correlation between all scenario pairs. Dimensionality is assessed by performing PCA across scenarios and quantifying how many components are needed to reach 90\% explained variance. Lower dimensional representations that maintain scenario separability are considered preferable, as they suggest the model has learned meaningful abstractions of the medical conditions.

\textbf{Task-specific validation.}
To validate that the assessment of entanglement had an impact on model performance, the pipeline tests the ability of a logistic regression model to perform pairwise decoding of feature identity. The dataset was split into a training and test set (0.8 and 0.2) with stratification by scenario type, and the Area Under the Receiver Operating Characteristic Curve (AUC-ROC) metric serves as our metric for model performance. We use logistic regression as a simple linear probe to assess the linear separability of features in the embedding space. High classification AUC-ROC values ($>$0.9) indicate that the embedding space preserves sufficient information for feature identification, suggesting effective disentanglement. Conversely, poor classification performance would imply that the embedded representations are entangled, necessitating further fine-tuning of the model to improve feature separability. The use of a simple linear classifier ensures that we are evaluating the quality of the representations themselves rather than the power of the classification model.

\begin{figure}[ht!]
\makebox[1.1\textwidth][c]{\includegraphics[width=1.2\textwidth]{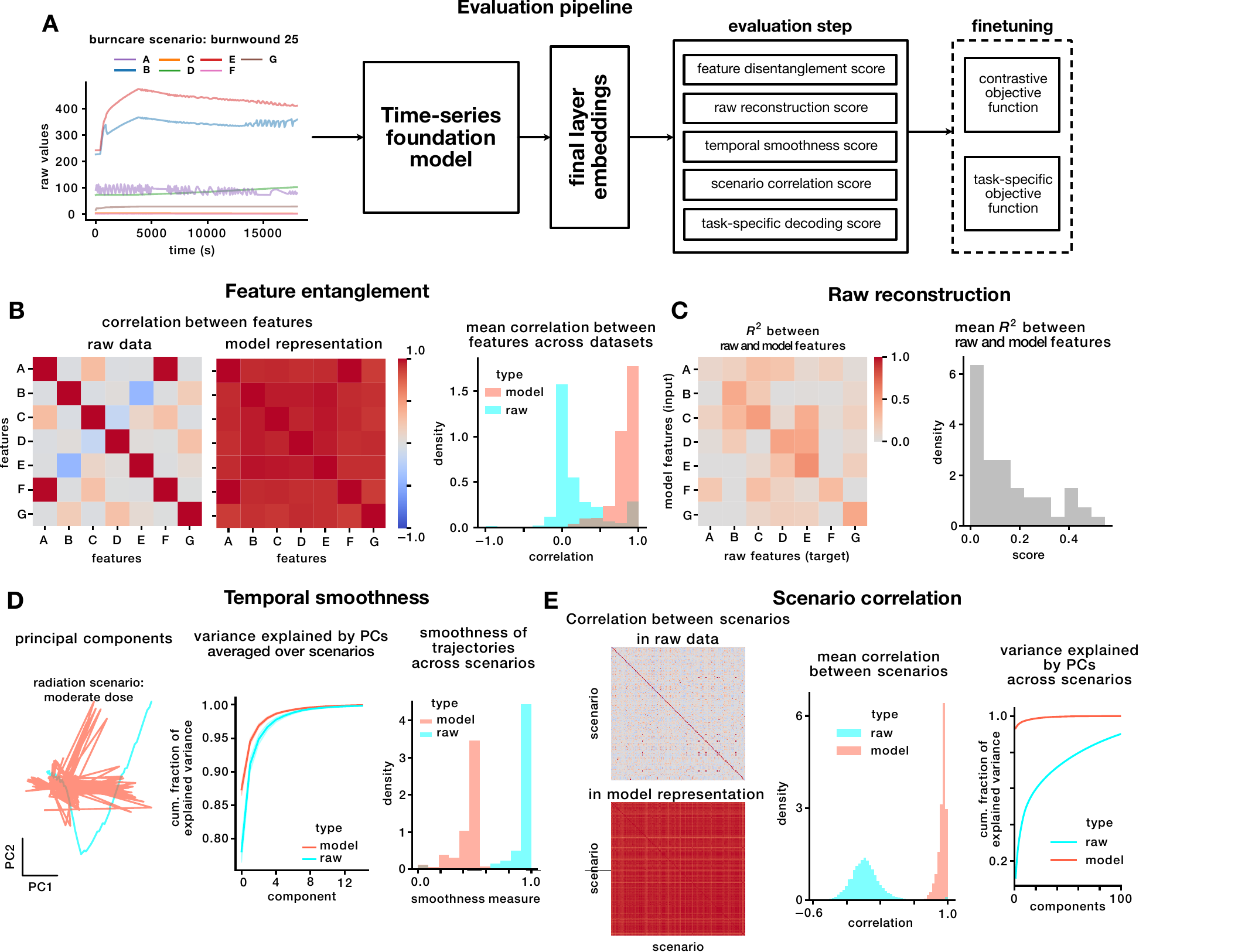}}
\label{fig:pipeline_diagram}
\caption{
\footnotesize{\textbf{Pipeline assessment of the Moirai Foundation Model.} Feature codings: A=arterial pressure (mmHg); B=carbon dioxide production rate (mL/min); C=central venous pressure (mmHg); D=heart rate (1/min); E=oxygen consumption rate (mL/min); F=renal blood flow (L/min); G=respiration rate (1/min). \textbf{(A)} Overview of the pipeline. \textbf{(B)} Correlations among features within a scenario for a single scenario of raw data (left), embedded features (middle) and the distribution of correlations across the entire matrix (right). \textbf{(C)} Correlation of features for raw and embedded data for a single scenario (left) and across scenarios (right). \textbf{(D)} Temporal dynamics of raw and embedded data for a single session (left), the mean variance captured by PCA across sessions (middle) and the smoothness of trajectories across sessions (right). \textbf{(E)} Correlation among scenarios for raw data of a single scenario (top left), embedded data of a single scenario (bottom left), across scenarios (middle), and the dimensionality of the dynamics (right).}
\vspace{-12pt}
} 
\end{figure}

\section{Results}
\subsection{Foundation Model Embeddings Introduce Systematic Distortions to Physiological Signals}

Our analysis revealed four major limitations in how the foundation model represents physiological time-series data, each suggesting specific challenges for medical applications. These limitations manifest as: (1) introduction of spurious correlations between features, (2) poor preservation of original signal characteristics, (3) loss of temporal dynamics, and (4) reduced ability to distinguish between different physiological features.

\subsection{Correlated noise is introduced through feature embeddings.}
While features in the raw physiological data showed minimal correlation, the features in the model's embeddings exhibited substantial correlation (Fig. \ref{fig:pipeline_diagram}B). This artificial correlation suggests that the model fails to maintain the independence of different physiological measurements, potentially confounding downstream medical analyses that rely on feature independence.

\subsection{Poor signal reconstruction from embeddings.}
Quantitative assessment revealed that the raw time-series signals were poorly reconstructed from their corresponding embeddings across all features and scenarios (Fig. \ref{fig:pipeline_diagram}C). This poor reconstruction indicates that essential physiological information is either lost or distorted during the embedding process.

\subsection{Destruction of temporal dynamics.}
Analysis of temporal structure revealed three key findings:
\begin{itemize}
    \item The PCA trajectories of raw signals showed smooth, physiologically plausible evolution over time, while embedded features exhibited erratic, discontinuous patterns (Fig. \ref{fig:pipeline_diagram}D, left).
    \item The embedded representations required significantly fewer principal components to explain 90\% of variance compared to raw signals, suggesting a concerning loss of signal complexity (Fig. \ref{fig:pipeline_diagram}D, middle).
    \item Trajectory smoothness metrics decreased across all scenarios (Fig. \ref{fig:pipeline_diagram}D, right).
\end{itemize}
These findings suggest that the foundation model fails to preserve the crucial temporal relationships that characterize physiological processes.

\subsection{Reduced dimensionality between scenarios compromises scenario discrimination.}
While the raw physiological signals maintained distinct characteristics across different medical scenarios, the embeddings showed substantially higher correlation between scenarios (Fig. \ref{fig:pipeline_diagram}E). This increased correlation was reflected in the reduced number of principal components needed to capture scenario variability (Fig. \ref{fig:pipeline_diagram}E).

\subsection{Reduced capacity for feature decoding impairs clinical utility.}
The ability to accurately identify distinct physiological features from the embedded representations was significantly compromised. While pairwise decoding of feature identity in the raw signals achieved near-perfect discrimination (mean AUC = 0.96 $\pm$ 0.05), the embedded representations showed markedly reduced performance (mean AUC = 0.78 $\pm$ 0.10). This degradation in feature discriminability was particularly pronounced for features that typically show subtle but clinically significant variations, such as central venous pressure and renal blood flow (AUC $<$ 0.75).

Collectively, these results demonstrate that the current foundation model architecture introduces systematic distortions that compromise the utility of the embeddings for medical applications. The combination of artificial correlations, lost temporal dynamics, and reduced feature discriminability suggests that significant architectural modifications or targeted fine-tuning will be necessary before these models can be reliably deployed in clinical settings. 

\section{Discussion}

We developed a rapid, resource-efficient pipeline for assessing the time series foundation models' capacity to generalize to problems in precision medicine. Our results suggest specific next-steps in pipeline development, and ways to integrate the pipeline into improved foundation model performance.  

\subsection{Next steps in pipeline development.}
Precision medicine fundamentally relies on understanding individual patient variability, yet our initial pipeline validation used generic simulations to assess model representations. This limitation presents several key areas for improvement:

First, we will expand our simulation scenarios to incorporate data-scarce medical situations that are critically important yet challenging to study in traditional clinical settings. Combat field trauma, for instance, represents an ideal test case due to its complex physiological cascades, time-sensitive nature, and limited real-world data availability. We plan to generate scenario variants that explicitly model patient-specific factors. For example, age-related physiological differences, particularly between pediatric and geriatric responses, represent a critical dimension of patient variability that must be incorporated into our simulations. These age-dependent variations interact closely with sex-specific differences in cardiovascular and metabolic parameters, which can significantly impact treatment outcomes. Furthermore, the complex interactions between comorbidities often determine treatment response patterns and must be carefully modeled. Of particular importance are genetic variations that influence drug metabolism and treatment efficacy, as these form the foundation of personalized therapeutic approaches.

Second, while our feature identity decoding provided initial validation of representation quality, the relationship between foundational model architecture and clinical performance requires deeper investigation. We propose incorporating additional validation tasks that directly assess clinical utility. Risk stratification using established clinical scores, such as NEWS2 and SOFA, will provide standardized metrics for model evaluation. The ability to predict treatment response trajectories will be crucial for clinical decision support, while early warning detection for clinical deterioration could enable proactive interventions. Additionally, accurate classification of underlying pathophysiological mechanisms will help ensure the model's representations align with medical understanding. These tasks will provide more nuanced insights into how model representations align with clinical decision-making requirements.

Finally, our initial focus on the Moiria model, while instructive, represents only a single point in the broader landscape of time series foundation models. By expanding our evaluation to include diverse architectural approaches \cite{liang_foundation_2024}, we can advance our understanding in several key areas. Through systematic comparison, we can identify which architectural elements best preserve physiological signal characteristics and determine how different pre-training strategies affect medical domain transfer. This comprehensive evaluation will enable us to develop targeted recommendations for model selection in clinical applications, ensuring that the most appropriate architectures are deployed for specific medical use cases.

\subsection{Targeted fine-tuning using distributions of simulated data.}
While our results showed that the Moiria model's base embeddings inadequately captured physiological signal characteristics, the pipeline offers a systematic approach to improving model performance. The key advantage of our simulation-based approach is the ability to generate targeted training data that addresses specific failure modes. Feature entanglement can be addressed by generating scenarios with controlled correlations between physiological parameters. This approach allows us to isolate and examine specific interactions, thereby improving the model's ability to disentangle complex physiological signals. Temporal dynamics can be enhanced by creating scenarios that emphasize clinically relevant timescales and trajectories, ensuring that the model captures the essential temporal patterns inherent in physiological processes. Additionally, scenario discrimination can be improved through the careful design of edge cases and boundary conditions, which help the model distinguish between different medical scenarios more effectively.

A critical next step is developing efficient fine-tuning strategies that leverage these synthetic datasets while ensuring generalization to real-world data. This involves investigating optimal mixing ratios between synthetic and real data during fine-tuning to balance model training effectively. Furthermore, developing validation metrics that specifically assess the reduction in identified failure modes will be essential for evaluating model improvements. Creating benchmark datasets that span common clinical scenarios and rare but critical edge cases will provide a comprehensive framework for testing and refining the model's capabilities.

The ultimate goal is to develop a systematic approach for improving foundation model performance in medical applications while maintaining computational efficiency and clinical relevance.

\begin{credits}
\subsubsection{\ackname} This study was supported by Mountain Biometrics, as well as the NIA's Artificial Intelligence and Technology Collaboratories for Aging Research program. Work was conducted outside the academic appointments of C.G., B.L. P.K. and W.W.P.

\subsubsection{\discintname}
Authors have a financial interest in Mountain Biometrics.
\end{credits}

\bibliographystyle{plain}
\bibliography{references}

\end{document}